\pdfoutput=1

\documentclass[11pt]{article}

\usepackage[final]{emnlp2021}

\usepackage{times}
\usepackage{latexsym}

\usepackage[T1]{fontenc}

\usepackage[utf8]{inputenc}

\usepackage{microtype}

\usepackage{graphicx}
\usepackage{tabularx}

\title{Automatic Construction of Enterprise Knowledge Base}

\author{Junyi Chai, Yujie He, Homa Hashemi, Bing Li, Daraksha Parveen, Ranganath Kondapally, Wenjin Xu\\
  Microsoft Corporation \\
  \texttt{juchai,yujh,hohashem,libi,daparvee,rakondap,wenjinxu@microsoft}}

\begin{document}
\maketitle
\begin{abstract}
In this paper, we present an automatic knowledge base construction system from large scale enterprise documents with minimal efforts of human intervention. In the design and deployment of such a knowledge mining system for enterprise, we faced several challenges including data distributional shift, performance evaluation, compliance requirements and other practical issues. We leveraged state-of-the-art deep learning models to extract information (named entities and definitions) at per document level, then further applied classical machine learning techniques to process global statistical information to improve the knowledge base. Experimental results are reported on actual enterprise documents. This system is currently serving as part of a Microsoft 365 service.
\end{abstract}

\section{Introduction}

Massive knowledge bases constructed from public web documents have been successful in enriching search engine results in Bing and Google for over a decade \citep{Noy2019}. There is growing interest in automatically constructing a similar knowledge base for each enterprise from their internal documents (e.g., web pages, reports, emails, presentation decks; textual contents in natural language form are all referred to as documents in this paper). Such knowledge base can help an enterprise to better organize its domain knowledge, help employees (users) better find and explore knowledge, and to encourage knowledge sharing. 

Mining knowledge from enterprise documents poses unique challenges. One challenge is that the system needs to be fully automated without per enterprise customization or existing (semi-) structured sources. Knowledge base construction from web documents is often based on bootstrapping entities from human-curated sources such as Wikipedia with customized extraction rules (DBpedia: \citealp{DBPedia07}, Freebase: \citealp{Bollacker2008}, YAGO2: \citealp{Hoffart2013}), or the existence of a prior knowledge base (Knowledge Vault: \citealp{Dong2014}). Maintaining such Wiki site and keep it fresh is costly for enterprise. Another challenge is that most training data for natural language processing (NLP) models is from public documents. Enterprise documents can have different writing style and vocabulary than the public documents. The data distributional shift \citep{DataShift2008} is a challenge as (a) we need model to generalize better to enterprise domain and (b) we need test metrics to reflect the actual performance on enterprise documents to guide model development. 

On the other hand, enterprise domain brings new opportunities. For search engines, the knowledge base must be extremely accurate. This requirement limits the usage of NLP models to extract information from unstructured text as few models can achieve the required precision with meaningful coverage. In enterprise domain, we can relax the requirement on accuracy as enterprise users are expected to spend more time to absorb and discriminate information. In addition, users can curate and improve the automatically constructed knowledge base, which is not an option for search engine users. The relaxation on accuracy requirement makes it possible to perform knowledge mining on unstructured text by heavily relying on NLP techniques.

In this paper, we present the first large-scale knowledge mining system for enterprise documents taking advantage of recent advances in NLP such as pretrained Transformers \citep{devlin-etal-2019-bert} as well as traditional NLP techniques. It is in production since February 2021 as part of a Microsoft 365 feature (Microsoft Viva Topics\footnote{\url{https://www.microsoft.com/en-us/microsoft-viva/topics/overview}}). For an enterprise that enables this feature, our system will build a knowledge base from its internal documents that already exist in Microsoft cloud and will keep it fresh without the need of any customized intervention. At the core of our knowledge base are entities mined from documents that are of interest to the enterprise, such as product, organization and project. These entities are loosely referred to as topics to the end users (not to be confused with topic modeling in NLP). The knowledge base is a collection of “topic cards” with rich information: properties that help users understand the topic (such as alternative names, descriptions, and related documents), or enable users to connect with people who might be knowledgeable about the topic (related people) or explore related topics.

The contributions of this work are as follows:
\begin{itemize}
\item We demonstrate a system in production that performs knowledge mining in large scale: hundreds of millions of documents, thousands of organizations.
\item We apply state-of-the-art deep learning models in two NLP tasks named entity recognition (NER) and definition extraction. We discuss the challenges and how we improve our system to reach the desired performance.
\end{itemize}

\section{System description}

\begin{figure*}[t]
\includegraphics[width=12cm]{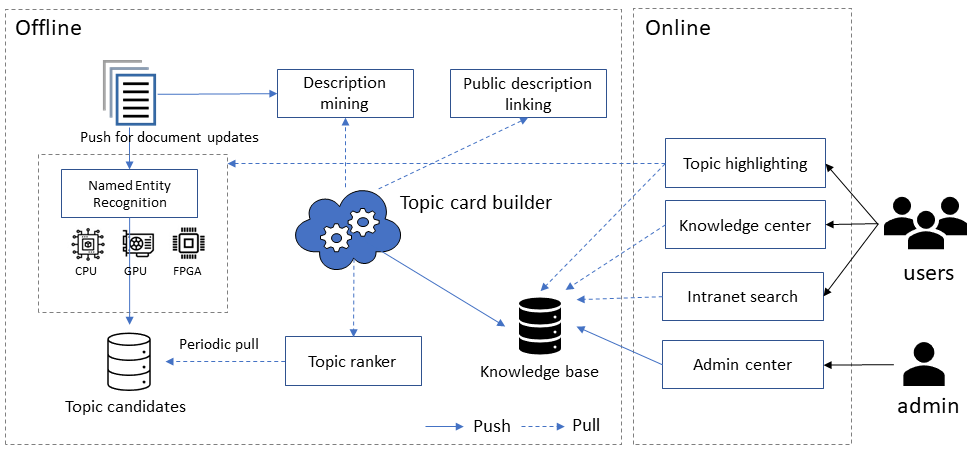}
\centering
\caption{An illustration of the knowledge mining system.}
\label{fig:system}
\end{figure*}

The overall system architecture is depicted in Figure \ref{fig:system}. In this section, we discuss at length the knowledge mining system that works “offline”. The system works in a semi-streaming mode: whenever there’s a document update, the content of the document is sent to the NER and description mining components. The NER component extracts entities then updates information in the topic candidate store. The topic ranker periodically pulls the topic candidates store to select the top $N$ topics. The topic card builder then builds topics cards with various attributes. Note that this is a simplified view of the actual system. For example, there is another component that conflates information from other sources using techniques described in \citet{winn2019alexandria}.

\subsection{Named entity recognition for enterprise}

NER is the typical first step in information extraction (\citealp{Jurafsky2009}, Chapter 22). Based on our study on enterprise customers’ demand and an analysis of Bing’s Satori knowledge graph, we define 8 entity types that are of interest to the enterprises while covering most of real-world entities. Among them, “person”, “organization”, “location”, and “product” are the common NER types in various public datasets (CoNLL03: \citealp{TjongKimSang2003}, OntoNotes: \citealp{hovy-etal-2006-ontonotes}; WNUT 2017: \citealp{Derczynski2017}), while “project”, “field of study”, “creative work” are less common but are also of high interest to enterprises. These 8 types cover about 85\% of entities in Bing’s Satori knowledge graph. The remaining entities are mainly biological organisms.

Our NER model is based on Transformers with the pretraining-finetuning paradigm \citep{devlin-etal-2019-bert}. State-of-the-art results on several NER benchmarks are achieved with Transformers \citep{Yamada2020,Li2020}. To make data collection easier, we train our model on public data but apply it to enterprise domain. The distributional shift between training and testing can cause a significant performance drop \citep{DataShift2008}. To measure model’s true performance under distributional shift, we construct a test set from actual internal documents within Microsoft. The size of this test set is comparable to CoNLL03 test set \citep{TjongKimSang2003}.

To mitigate the distributional shift issue, we divide model training into multiple stages, with the first stage training on large amount of automatically annotated data using Wikipedia, which has been shown to help the system generalize better to a new domain \citep{Ni2016}. Entities are identified by wikilink, and we use Bing’s Satori knowledge graph to find out the corresponding entity type. We selected paragraphs with at least 10 wikilinks, which gives us $\sim$1 million paragraphs. Finally, we use an entity linking tool NEMO \citep{Cucerzan2007,Cucerzan2014} to annotate entities without wikilinks and get $\sim 50\%$ more entities.

The benefit of Wikipedia training data lies in its size, but it comes with low annotation quality. After training on it, we continue training on smaller data with high quality human annotation. In the second stage, we use OntoNotes 5.0\footnote{\url{https://catalog.ldc.upenn.edu/LDC2013T19}} data set and mapped their types to our 8 types. This stage is mainly beneficial for the common NER types, but it does not help our additional “project” and “field of study” type. In the last stage, the training data is a combination of a small number of web documents with 8 types annotation (size is $\sim$1000 paragraphs) and CoNLL03 data with “MISC” type being reannotated to one of our 8 types. This last stage of training data is most aligned with our NER type definition.

To illustrate the effect of multistage training as well as additional improvement techniques, we consider BERT-base with cased vocabulary finetuned only on the last stage of training data as a strong baseline. The F1 metrics of our model, baseline, and ablation experiments on our test set from internal documents are shown in Table 1. The baseline 56\% F1 is much lower than the reported F1 > 90\% on CoNLL03 test set \citep{devlin-etal-2019-bert}, which shows the challenge from the distributional shift (we also test our baseline model on CoNLL03 test set and get F1 > 90\%). The most common entity type in our test set is product, which can be more difficult to detect than the most common entity types (person, location, organization) in CoNLL03. Also our test set is noisier than CoNLL03 as internal documents are often less formal than public documents such as newswire articles. Our best model achieved an F1 of 71.1\%. In ablation study, multistage training improves F1 by 5.4\% from the baseline. We find additional techniques that can robustly improve model performance:

\begin{figure}[t]
\includegraphics[width=7.5cm]{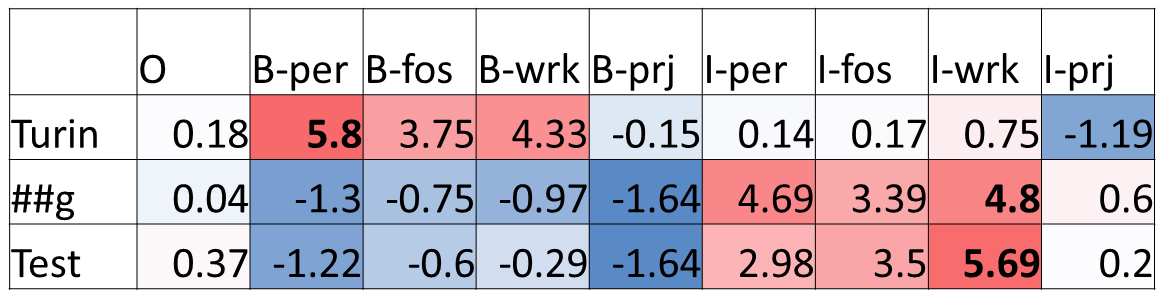}
\centering
\caption{Scores for selected tokens and selected types from the sentence “The history of NLP dates back to the 1950s when Alan Turing proposed a simple test (the “Turing Test”) to determine …”. The abbreviations in use are: per for person, fos for field of study, wrk for creative work, and prj for project.}
\label{fig:viterbi}
\end{figure}

\begin{itemize}
\item \emph{Data augmentation}: we find two most useful data augmentation methods out of many methods we have tried. One method is simply lower casing the training data. This method has been shown to increase NER performance on uncased text significantly, and even improve performance on cased text when train and test on different domains \citep{Mayhew2019}. The second method is to replace an entity mention with a randomly selected entity of the same type. This is motivated by our observation that the distribution of entities roughly follows the Zipf’s law. Randomly replacing entities can give more weights to tail entities. In combination, data augmentation provides a 0.6\% F1 lift.

\item \emph{Focal loss}: NER is an imbalanced classification problem as most input tokens are not entities. We test loss functions suitable for imbalanced dataset: Dice Loss \citep{Li2020}, Am-Softmax \citep{Wang2018}, and Focal Loss \citep{FocalLoss2017}. They all provide similar improvement. We report focal loss (hyperparameter gamma=1.6) results here with an additional 1.5\% F1 lift.

\item \emph{Viterbi decoding}: as there is no hard constraint on the sequence of labels from BERT, the sequence can be invalid under the standard BIO tagging scheme \citep{Lample2016}. Figure \ref{fig:viterbi} shows such an example. The scores for tokens (“Turin”, “\#\#g”, “Test”) give an invalid label sequence of B-per, I-wrk, I-wrk (per stands for person, wrk for creative work) using greedy argmax decoding, which we correct to O, O, O in the baseline setting. We observe that the correct sequence B-wrk, I-wrk, I-wrk has highest sum of scores among all valid sequences, for example B-per, I-per, I-per. Based on this observation, we use Viterbi algorithm to find the valid path under BIO scheme with the maximum sum of scores. This provides a 1.9\% F1 lift. We have also tried adding a CRF layer on top of BERT, training jointly or separately. We do not see additional gain (though CRF layer can further improve 1-layer student model).

\item \emph{Bigger and better pretrained model}: BERT is pretrained on English Wikipedia and BookCorpus, which have limited writing styles. Enterprise documents can be more diverse, less formal and noisier. Therefore, pretraining on more diverse corpora may help our task. We switched from BERT base-cased to UniLM v2 large-cased, which is pretrained on additional ~144GB corpora including OpenWebText, CC-News, and Stories \citep{pmlr-v119-bao20a}. This provides a 5.7\% F1 lift.
\end{itemize}

\begin{table}[t]
\resizebox{8cm}{!}{
\begin{tabular}{lllll}
\hline
Experiment                & Config                        & F1     & P      & R      \\
\hline
Best model               & UniLMv2-large: all techniques & 71.1\% & 72.2\% & 70.2\% \\
Baseline                  & BERT base: single last stage  & 56.0\% & 54.0\% & 58.2\% \\
Ablation & BERT base: multi-stage        & 61.4\% & 60.7\% & 62.1\% \\
                          & +data augmentation            & 62.0\% & 60.6\% & 63.4\% \\
                          & +focal loss                   & 63.5\% & 62.7\% & 64.4\% \\
                          & +Viterbi decoding                      & 65.4\% & 65.8\% & 65.1\%\\
\hline
\end{tabular}
}
\caption{NER results on internal test set.}
\label{tbl:ner}
\end{table}

For production, we distill knowledge from the 24-layer UniLMv2 teacher model into a 3-layer student model, which is initialized from the weights of the first 3 layers of the teacher model \citep{Hinton15}. We use 1 GB Wikipedia data for distillation. The student model suffers a 5.6\% F1 drop. Though not used in production, we experiment continuing distillation with only about 50MB of internal documents. This small amount of data reduces the gap between student and teacher models to 0.9\% F1, which suggests the usefulness of using in domain data for knowledge distillation.

Knowledge distillation gives us $\sim$6x speed up for inferencing a single input sequence on Nvidia V100 GPU with f32 precision. On top of student model, we get another $\sim$14x speedup by (1) exporting model from Pytorch to ONNX \citep{OnnxMsft}, (2) switching from Python to C\#, and (3) running inference in f16 precision and batch mode.

\subsection{Topic ranker}
In the NER step, tens of millions of topic candidates could be detected. The goal of topic ranker is to pick the top tens of thousands most salient topics while reducing the number of noisy topics. We achieve this in two stages by first simply ranking topics by their total number of times being detected by NER (referred to as NER frequency) to produce a short list of topics. Then we rerank the short list by scores from a binary classifier. The classifier is trained to distinguish between good and noisy topics. It uses features such as NER frequency, document frequency, topic-in-title frequency (number of times the topic appears in the document title) and the ratios of these counting features.

This classifier is effective as it uses global statistical information not available during NER stage. For example, the word “Company” could be mislabeled as an organization by the NER model. Although the probability is small, it could still make into the short list as this word appears very often. The classifier would filter it out as the ratio (NER frequency/document frequency) is very small.

Our training set contains 6000 annotated topics detected from about 0.5 million Microsoft internal documents. Using a single feature NER frequency as a baseline, the AUC is 0.54. We train a gradient boosting trees classifier \citep{NIPS2017_LightGBM} using 5-fold cross validation and achieve an average validation AUC 0.67.

In the production system, as the number of topic candidates scales up, the topic ranker could play a more important role as much lower percentage of topics will be selected. To evaluate its true usefulness, we apply the classifier in the end-to-end system to process all Microsoft documents. We randomly sample a subset of topics before and after applying the classifier. We observe a 9\% reduction in noisy topics with the classifier.

\subsection{Definition mining}

A succinct and accurate description is a crucial attribute of a topic. Such descriptions come from two sources: (1) for some topics such as “field of study” type, their descriptions exist in public knowledge and therefore we retrieve this information from Bing’s Satori knowledge graph using an existing context aware ranking service, which is in use for Microsoft Word’s Smart Lookup feature; (2) more importantly, we build a description mining pipeline to extract private definitions from enterprise documents. This pipeline consists of the following steps:

\begin{enumerate}
\item Split a document into sentences.
\item A deep learning model classifies each sentence into one of 5 categories. Pass a sentence in the “Sufficient Definition” category to the next step.
\item Extract topic from the sentence using a list of patterns, for example: {topic} is defined as {description text}.
\item Remove sentences with negative opinion (or sentiment) based on lexicon match. We use the Opinion Lexicon from \citet{Hu2004}.
\end{enumerate}

A large corpus contains definition-like sentences with a wide range of ambiguity beyond a binary classification task can capture. Therefore, we make the task more granular and define 5 categories most common in enterprise domain: Sufficient, Informational, Referential, Personal and Non- definitions. Detailed schema is included in the Appendix.

To collect training data, we need to first collect sentences with a relative high chance of being a description. In addition, we want to collect more hard negative examples such as opinions (e.g., "Caterpillar 797B is the biggest car I've ever seen.") than easy negative examples. Using query log from Bing, we achieved these two goals: we collect search results for queries that match patterns such as “what is \{term\}”, “define \{term\}” as the results are highly related to definitions. The search results also have the advantage of being more diverse than a single corpus. From the search results, we create a set of 42,256 annotated sentences, which is referred as public dataset. As we will show, a model trained on the public dataset suffers a significant performance degrade on enterprise documents due to distributional shift. Therefore, we construct a second dataset from our internal documents that have been approved for use after compliance review, which is referred as enterprise dataset. The model trained on the public dataset is used for identifying candidate sentences for annotation during the construction of the enterprise dataset. Using the enterprise dataset involves many compliance restrictions. For example, we need to delete a sentence if its source document is deleted or our access expires; the model is trained within the compliance zone and stays within it. Details for these two datasets are shown in Table \ref{tbl:defdata}, which also includes the DEFT corpus for comparison \citep{Spala2019}. Roughly 15\% of the data from the two datasets is withhold from training for testing.

\begin{table}[t]
\centering
\resizebox{7.5cm}{!}{%
\begin{tabular}{lll}
\hline
Dataset & \# of sentences & \# of positive \\
\hline
Public dataset & 42,256 & 10,927 \\
Enterprise dataset & 58,780 & 49,017 \\
DEFT (Spala et. al. 2019) & 23,746 & 11,004 \\
\hline
\end{tabular}%
}
\caption{Datasets for definition classification task.}
\label{tbl:defdata}
\end{table}

Similar to our approach in NER, we consider BERT-base (with cased vocabulary) as a strong baseline. First we train BERT-base model on the public dataset. When testing it on public and enterprise datasets, we get F1 results of 0.82 and 0.64 respectively, as shown in Table \ref{tbl:defResult}. This performance degradation again exemplifies the challenge from distributional shift. Then we train on the enterprise dataset and compare BERT-base with BERT-large and UniLMv2-large. UniLMv2-large achieves the best result with F1 of 0.75, which may again benefit from the bigger pretraining corpus \citep{pmlr-v119-bao20a}. In Table \ref{tbl:defResult}, we also add the result from rule-based classification, which directly uses the list of patterns in Step 3 (e.g., “is a”, “is defined as”, “refer to”) to identify definition. It is evaluated as a binary classification task: “Sufficient Definition” vs Others. We get F1 of 0.48 with an even lower precision of 0.40. This shows the necessity of model-based classification in Step 2 in our definition extraction pipeline.

\begin{table}[t]
\centering
\resizebox{7.5cm}{!}{
\begin{tabular}{llll}
\hline
Model & Train data & Test data & F1/P/R \\ \hline
{Bert-base} & {Public} & Public & 0.82/0.76/0.89 \\
 &  & Enterprise & 0.64/0.55/0.77 \\
\hline
BERT-base & {Enterprise} & {Enterprise} & 0.73/0.68/0.80 \\
BERT-large &  &  & 0.72/0.70/0.77 \\
UniMLv2-large &  &  & 0.75/0.71/0.80 \\ \hline
Rule based & N/A & Public & 0.48/0.40/0.60 \\ \hline
\end{tabular}
}
\caption{Results for definition classification.}
\label{tbl:defResult}
\end{table}

For production, we distill our best model into a much smaller BiLSTM model. The embedding of the BiLSTM is initialized from 50-dimensional Glove vector \citep{Pennington2014} with a reduced vocabulary size of 0.12 million. The hidden dimension size is 300. We follow similar knowledge distillation approach as in \citet{Tang2019}. The student model reaches F1 of 0.72 while achieves about 30x speedup vs. the 24-layer teacher.

\subsection{Topic card builder}

Topic card builder builds topics cards with rich information by aggregating information like definition and acronym from other components. More importantly, it computes the relatedness between topics, documents and users. Using relatedness, it links the top related topics, documents, and users to each topic. By adding related users to a topic, we enable the “expert finding” scenario, which is important for enterprise users to explore expertise. Topic card builder also conflates two topics if their degree of relatedness exceeds a threshold and they pass additional checks to prevent over conflation.

To compute relatedness between any two items, we build a dense embedding vector for each topic, document and user. We apply SVD to decompose the topic-document matrix $M$ into topic and document embeddings, where $M_{i,j}$ is the BM25 of topic $i$ in document $j$. This is a classical algorithm in collaborating filtering \citep{Koren2009} and semantic embedding \citep{Bellegarda2000,Levy2015}, but the challenge is the size of the matrix $M$ in the $j$ dimension as it can be on the order of tens of millions. We improve a randomized SVD algorithm that iterates on smaller batches of documents so it can solve problem of our scale on a single machine under 8 GB memory limit \citep{Halko2011}. User embedding is represented as the average of embeddings of the documents that the user has authored. Relatedness is computed as the dot product of two embedding vectors. Top $K$ topics and users most related to a given topic are added to the topic card in this manner. For related documents, embedding is used as a recall-oriented step to select candidate documents, and we apply an additional reranking step using additional signals.

To evaluate the overall quality of the system, we conduct human evaluation on the quality of generated topic cards (70K) mined from Microsoft internal documents (40 million). We ask users (Microsoft employees) to judge the overall quality of randomly sub-sampled topic cards by considering all the information. A good topic card means that it has sufficient information to help users understand the topic. In this study, we achieve 85\% good rate.

\section{Use Cases}

\begin{figure}[t]
\includegraphics[width=5cm]{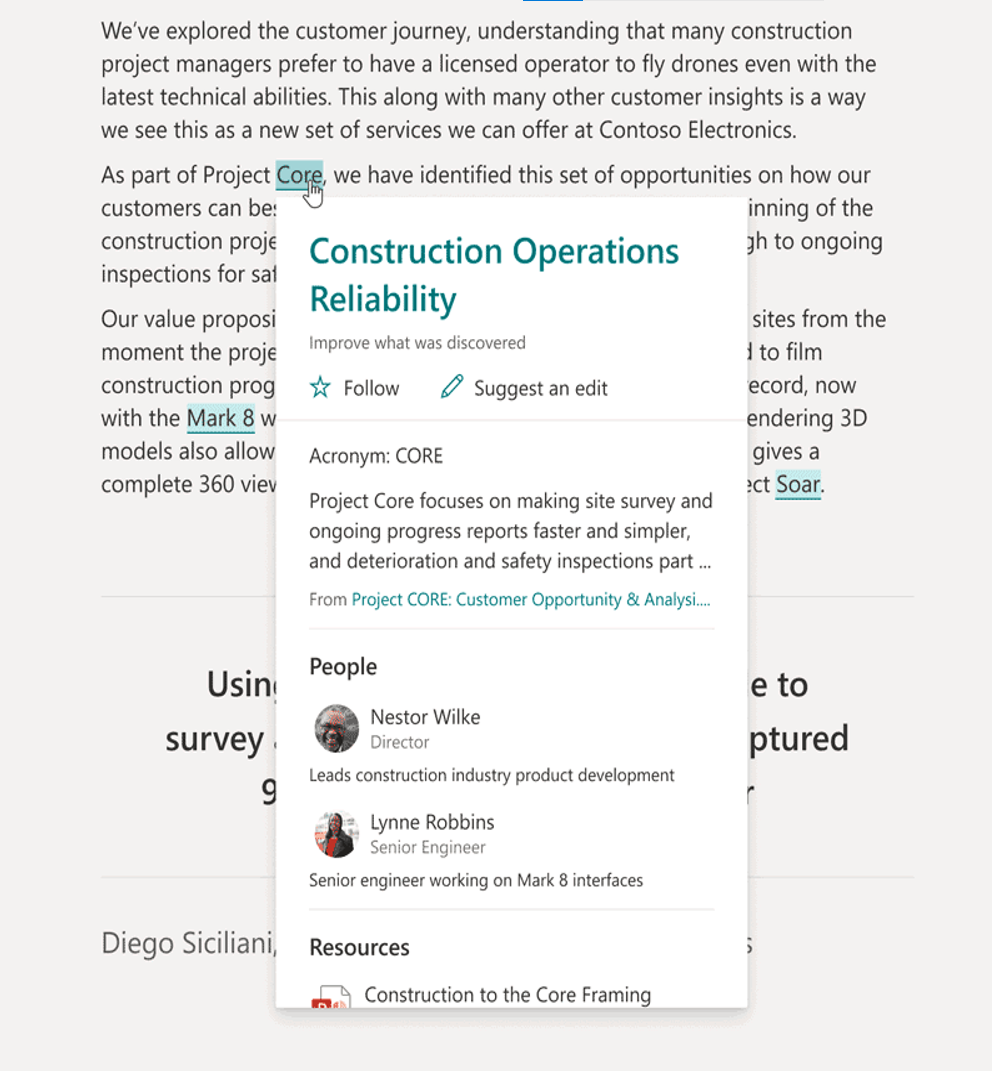}
\centering
\caption{Snapshot of an example topic card impression in enterprise web document.\footnotemark[4]}
\label{fig:topicHighlight}
\end{figure}

\begin{figure}[t]
\includegraphics[width=7.5cm]{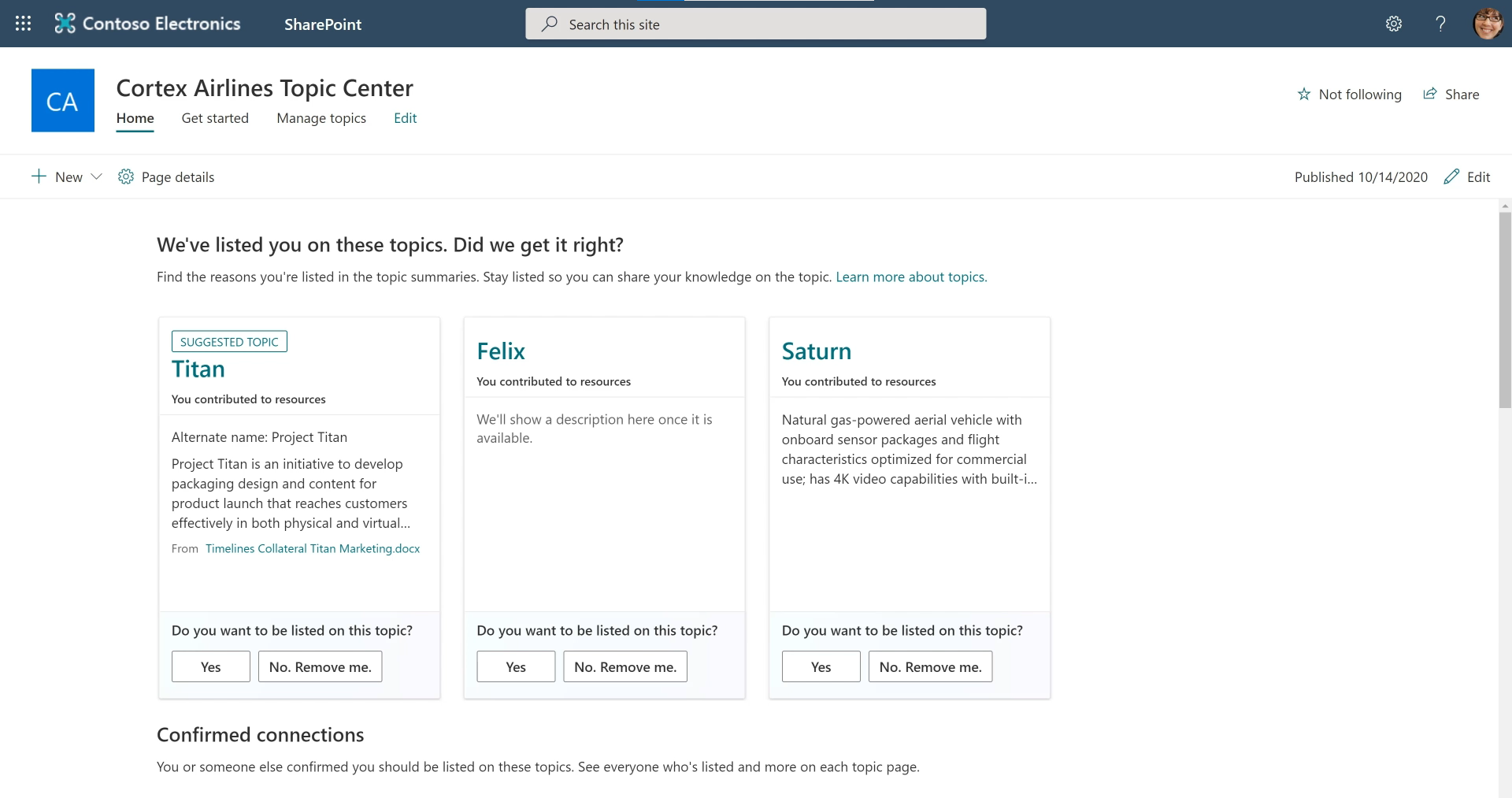}
\centering
\caption{Snapshot of a personalized topic center homepage.\footnotemark[4]}
\label{fig:topicCenter}
\end{figure}

The detailed user guide and licensing information can be found on Microsoft Viva Topics website\footnote{\url{https://docs.microsoft.com/en-us/microsoft-365/knowledge/}}. Here we briefly introduce two ways user can interact with the knowledge base. Figure \ref{fig:topicHighlight} shows the topic highlighting feature. Topic mentions in documents get automatically linked to the knowledge base. User can hover over the topic mention to see the topic card and click the link in the topic cards to checkout more information. Figure \ref{fig:topicCenter} shows an example topic center homepage. The view is personalized as the related topics for a user is presented for the user to confirm. Users can also checkout all the topics from Manage Topics page.

\footnotetext[4]{The contents (company name, topic information) are not real internal information but are created for demo purpose.}

\section{Conclusion}
Organizing resources inside an enterprise into one centralized location facilitates knowledge and expertise sharing and improves productivity. We present a system that automatically constructs a knowledge base from unstructured documents to help enterprises achieving this goal. The system is built upon a combination of deep learning models and classical techniques. We show the challenge of applying NLP models in enterprise domain. We also discuss how we improve the models and the whole system to meet our requirements with detailed experiment results. Finally, we show two typical use cases. We hope our experience can benefit researchers and practitioners in this field.

\bibliography{anthology,custom}
\bibliographystyle{acl_natbib}

\appendix
\section{Appendix}

\begin{table*}[t]
\centering
\begin{tabular}{p{2cm}p{5cm}p{7cm}}
\hline
Category & Description & Example \\
\hline
Sufficient Definition & Can uniquely define and can only define this term. & Statistics is a branch of mathematics dealing with data collection, organization, analysis, interpretation, and presentation. \\ \hline
Informational Definition & Informational but cannot uniquely define this term or can also apply to other terms. & Statistics is a branch of mathematics. \\ \hline
Referential Definition & Is a definition but does not contain the term but instead a reference (“it/this/that”). & This method is used to identifying a hyperplane which separates a positive class from the negative class. \\ \hline
Personal Definition & Associated with the name of a person. & Tom is a Data Scientist at Acme Corporation working on natural language processing. \\ \hline
Non-definition & Does not fall in the other categories. It can be an opinion (hard negative) or not related to definition at all (easy negative). & The Caterpillar 797B is the biggest car I've ever seen.\linebreak "Effective Date" means the date 5th May 2020. \\
\hline
\end{tabular}%
\caption{Schema for definition categories.}
\label{tbl:defschema}
\end{table*}

\end{document}